# Boosting performance of computer vision applications through embedded GPUs on the edge

Fábio Diniz Rossi*
*Federal Institute of Education, Science, and Technology Farroupilha (IFFar) - Alegrete - Brazil
Email: {fabio.rossi@iffarroupilha.edu.br

*Abstract*—Computer vision applications, especially those using augmented reality technology, are becoming quite popular in mobile devices. However, this type of application is known as presenting significant demands regarding resources. In order to enable its utilization in devices with more modest resources, edge computing can be used to offload certain high intensive tasks. Still, edge computing is usually composed of devices with limited capacity, which may impact in users quality of experience when using computer vision applications. This work proposes the use of embedded devices with graphics processing units (GPUs) to overcome such limitation. Experiments performed shown that GPUs can attain a performance gain of up to 820.36% when compared to using only CPUs, which guarantee a better experience to users using such kind of application.

*Keywords*—Big data, computer vision, embedded platform, GPU.

## I. INTRODUCTION

Cloud computing [1] offers an elastic service-oriented architecture that is virtually unlimited in terms of resources and can process large amounts of data. The exponential growth of networked devices sending data to be processed by clouds in real time fostered the development of Big Data platforms [2]. However, the communication channels have become saturated due to the large amount of data that must be processed by the cloud environment. In order to address this limitation, edge computing brings computing resources closer to the customer, speeding up Big Data processing.

Edge devices [3] are usually embedded systems that perform pre-processing and real-time analysis at intermediate points of the network on the data coming from the most diverse equipment and sensors distributed in the environment. However, the embedded devices are developed with special-purpose and designed with strict issues of energy consumption. Therefore, such devices are resource-limited, especially regarding the number of processor cores, which can cause a loss in terms of quality of service as the amount of data to be processed increases.

One of the computing areas most sensitive to quality of service and processing capacity is computer vision [4]. Pattern detection, segmentation, and tracking are quite common operations in computer vision, which is concerned with using mathematical models on large data arrays. Architectures that keep GPUs, besides to offer a large number of specialized processing units for such calculations, preserve large areas of memory that can hold a more significant amount of data to be processed.

For this reason, the replacement of edge devices that use CPUs by embedded devices with GPUs is a solution that allows accelerating the processing of such applications. It occurs because massive amounts of GPUs can be placed in the same hardware design space that previously supported only CPUs. Therefore, **this paper proposes the use of edge devices with embedded GPUs in order to accelerate the processing of CV applications.** Results obtained in several experiments confirm the benefits of using such device in terms of performance gains, leading to an improved user experience.

The rest of the paper is organized as follows: Section II presents a theoretical reference on computer vision and related work. Section III presents the scenario, applications, evaluations, and discussion of results. Finally, Section IV presents some conclusions and future work.

## II. BACKGROUND AND RELATED WORK

In Big Data environments [2], cloud providers offer data analysis and integration services in the form of an Analytics-as-a-Service (AaaS). The exponential increase in the number of sensors generating data to be processed by such services stimulated the advent of the Internet-of-Things (IoT) [5]. To meet this growing demand in the amount of data to be transmitted between the customer and the cloud, it is necessary to increase the capacity of the communication channels. However, the increase in this type of infrastructure is expensive. For a time, techniques such as offloading [6] tried to be a palliative to solve the limitations of latency in the data transfer. Currently, edge computing [3] is a low-cost change in infrastructure that addresses such limitations, bringing cloud services closer to the customer and offering embedded devices with moderate processing power.

One area of computer science that requires more computational resources due to the quantity and quality of data coming from sensors is CV. Bahl et al. [7] cites CV as one of the challenges to be met by cloud environments, precisely because of the expressive amount of data that must be sent over the network. Besides, Abdeslam et al. [8] analyze the rise in execution time when increasing the amount of data to be processed, even when using optimized CV techniques. White et al. [9] proposes to analyze the data offline, in a high performance cluster. However, CV often requires real-time pattern identification. Ashok et al. [10] presents a vehicle monitoring system through cameras and the authors propose to send such data for analysis in a cloud environment. The work reinforces what was discussed earlier when it presents offloading as an option to address the limitations of network latency in sending data to be processed in cloud environments.

When we consider the processing of CV applications over GPUs, its use is still more common in large-scale environments such as clusters and clouds. Campos et al. [11] offers a high-performance computing environment with GPUs, but falls within the two limitations presented above: high latency in data transfer to the cluster and offline processing of data.

To the best of our knowledge, no other work has brought real-time analysis of CV applications to edge devices equipped with GPUs, which can accelerate data traffic by being positioned closer to the users, in addition to allowing improved performance of the data analysis due to the availability of graphics processors.

## III. EVALUATION AND DISCUSSION

Computer vision covers a broad range of applications in the context of edge devices, from accident detection in highways [12] to recognition of fugitives through analysis of images of security cameras [13]. In this sense, we conducted experiments to evaluate the impact of using GPUs in edge devices during the execution of different CV algorithms. The algorithms are:

- Haar Cascade [14]: It is an algorithm for detecting objects. In general, this algorithm is used for detecting pedestrians, car accidents, and even more specific events such as particular facial expressions. Haar Feature-based Cascade Classifiers receive as input two types of sequences of images: the elements that need to be found (e.g., faces of different persons, cars crashing, etc.) and other random items that can include chairs, tables, and so on. Next, the algorithm analyzes the inputs looking for common characteristics of the elements that need to be found, and what are the differences between them and the random items through classification techniques. Once the analysis is finished, the classifier algorithm is ready to be used for detecting patterns in real-life scenarios .

- Generalized Hough [15]: It was developed based on the Hough transformation algorithm which aims to identify defined types of shapes, such as: lines, circles, and ellipses. However, Generalized Hough can be used to detect arbitrary shapes (e.g., shapes having no analytic form). It uses edge information to define a mapping of objects from the orientation of an edge point to a reference point of the shape.

- Hough Lines [16]: In order to perform object recognition, it is critical to reduce the image size while preserving its main characteristics and structural information. In this sense, the determination of the position and orientation of straight lines in images is of great importance in the fields of computer vision and image processing. Hough Lines uses a technique applied in image analysis to find instances of objects within a particular class of shapes identifying its edge lines. Initially it was concerned only with the identification of lines in the image, but later has been extended to identify positions of arbitrary shapes.

- Hog [17]: Histogram of Oriented Gradients (HOG) aims to identify objects in an image by analyzing the distribution of intensity gradients. In this sense, the Hog features are extracted based on the geometric properties of the object. These properties are used in many applications, such as: hand gesture recognition, traffic sign recognition, human recognition, among others.

- Super-Resolution [18]: It is an algorithm developed to perform techniques that enhance the resolution of an image. This algorithm explores a sharpness index in order to optimize the low-resolution images to high-resolution.

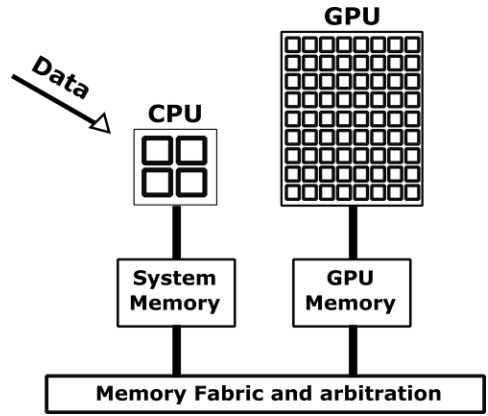

Fig. 1: Jetson TX2 architecture. Data is acquired by CPUs and stored in system memory. In order to use the GPUs, data must be transferred from system memory to GPU memory.

In order to evaluate the CV algorithms performance on GPUs, we have adopted the embedded System-on-a-Chip (SoC) development kit Jetson TX2[1]. This board is suitable for high performance computing applications, such as robots, drones, smart cameras, and portable medical devices. In addition, this embedded device comprises a GPU with 256 CUDA cores making it ideal for our experiment. Figure 1 shows an overview of the Jetson TX2 architecture. Tests on the GPU platform were compared to a SoC using ARM Cortex-A57 quad-core processor, a general-purpose platform used for edge computing. Linux Ubuntu 16.04.3 LTS (kernel 4.4.15-tegra) was used in all experiments on both platforms.

To perform the experiments we used implementations of the chosen algorithms in OpenCV (Open Source Computer Vision Library) version 3.3.0 [19]. OpenCV is an open source computer vision and machine learning software library. This library provides a comprehensive set of these applications. Besides, OpenCV is compatible with CUDA (Compute Unified Device Architecture) [20], which is a parallel computing platform and application programming interface. CUDA allows coding in C, C++, and Fortran directly to the GPU, producing a significant increase in performance during the execution of several types of applications by using the power of GPUs. To validate the experiments, the results presented below are the mean of 10 executions of each algorithm with a standard deviation less than 2%.

---
[1]Available at: <https://developer.nvidia.com/embedded/buy/jetson-tx2>.

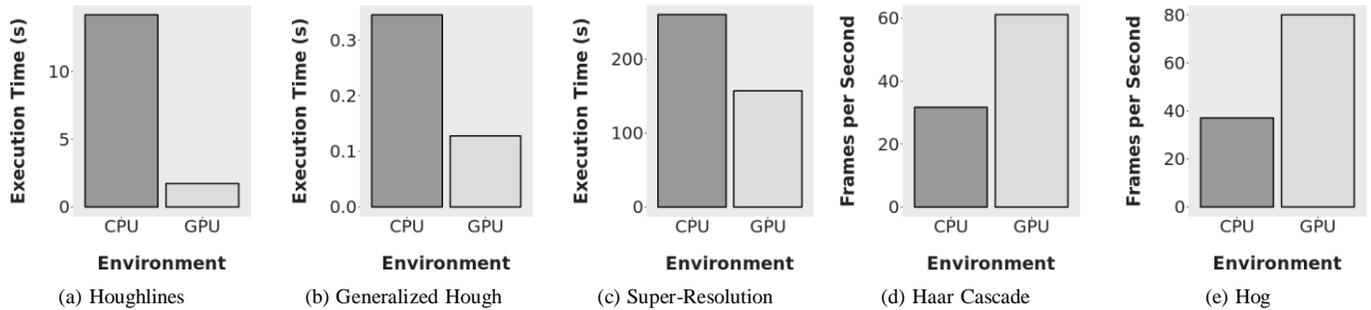

Fig. 2: Performance comparison between CPU and GPU during the execution of different computer vision algorithms in an edge device.

The results, presented in Figure 2, revealed that the GPU utilization provided performance gains when executing all the chosen CV applications. The most significant gain was achieved by the algorithm that uses Hough transform to detect shapes in images. Running such algorithm on the GPU increased the performance in 820.38%. Such positive result can be understood by analyzing the structure of such algorithm. Its implementation allows the exploitation of the massive parallelism provided by the GPU.

The results also showed that using GPU instead of CPU to execute template matching algorithms based on Generalized Hough Transform (Figure 2b) could generate performance gains of up to 270.54% since such type of work involves several steps that can be distributed among the high number of processing units in the GPU. For example, this algorithm manipulates information related to the edge points of the tested image into matrices. At this point, the GPU approach shows a considerable advantage over the CPU by processing greater amounts of matrices data at the same time into their cores.

The Hog feature descriptor algorithm also presented positive results by running on the GPU (gain of 215.72%). This performance gain is explained through the analysis of the algorithm behavior. The Hog descriptor calculates the gradient orientations and magnitude of each pixel of an image. Next, it divides the image into groups of pixels (which are called cells). It creates histograms of gradient orientations of each of these cells and groups them into blocks that will be analyzed by a descriptor. This algorithm takes advantage of the parallelism provided by the GPUs since each of these tasks can be divided into small pieces and performed simultaneously. Therefore, the Hog algorithm distributes several threads among the GPU cores to process multiple pixels, cells, and blocks at the same time.

Running the Super-Resolution algorithm upon the GPU resulted in a performance gain of 165.59%. Despite the better performance, it can not be considered so expressive when compared with the algorithms presented before. This result can be understood by analyzing in details the Super-Resolution algorithms' behavior. First, it analyzes each image sequence, detecting points with poor quality and replacing such points by others compatible ones from the other frame that presents better quality. The search by similar points can include the analysis of previous or even upcoming image sequences. In this context, depending on the image sequence that is being processed, Super-Resolution algorithms may have to look for new frames to access several different positions in the system memory. This irregular memory access pattern increases the chances of occurring performance-degrading events due to more cache misses.

The results showed that running the face detection algorithm with Haar Feature-based Cascade Classifiers on the GPU could achieve a performance gain of 93.36%. This smaller gain (when compared to the other algorithms) occurred due to a bottleneck caused by excessive data transfer among the memories since this algorithm needs to move the image sequence from the system memory to the GPU memory before starting the detection process. The results showed that using GPU could score significant performance gains during the execution of different CV algorithms on edge devices.

The results showed that performance gains can vary significantly (from 93.36% to 820.38%) according to the algorithms' behavior. The main reason for the considerable difference is the algorithm's capability of performing its tasks without requiring excessive data transfer between the system memory and the GPU memory. For example, the best result was achieved by the application that uses Hough transform for detecting shapes in images. This algorithm does not require data transfer among the GPU memory and the system's memory during its tasks. By having such behavior, the Hough transform shape detection algorithm achieved better results than the other algorithms that require considerable data transfer from the system's memory to the GPU memory.

The Hough transform shape detection algorithm (which presented the most significant gain when running on the GPUs) is composed of five phases. First, it recognizes the edges in the selected image using mechanisms that run entirely on the GPU. After, it analyzes the edge points that belong to each line of the image to detect all possible lines passing through it. Such phase takes a considerable advantage of GPU processing power since it does not require communication with the system memory. In a first moment, each thread converts a part of the image to an array of pixel coordinates in the GPU's shared memory. Then, a second thread processes the array of pixel coordinates to create a Hough line also in the GPU's shared memory. Only when the threads finish processing the entire set of coordinates, the Hough line is copied to the corresponding

Hough space in the system's memory.

On the contrary, the face detection algorithm with Haar Feature-based Cascade Classifiers requires intensive data transfer between the system's memory and the GPU's memory. As a result, it presents a bottleneck caused by the data transfer itself and performance-degrading events such as cache misses that are caused when the running application frequently requires access to different positions in memory. Therefore, this algorithm was not able to achieve more significant gains despite the massive level of parallelism provided by the GPU. The results show that there is an inability of the algorithm to use the power of the parallelism provided by GPUs and the memories coherence.

## IV. Conclusions and Future Work

The exponential growth of mobile devices and the increase in data exchange, storage and processing are leveraging the research in Big Data. In order to attend the processing demands of Big Data in this scenario, edge computing emerged bringing cloud capabilities closer to the customer. Edge computing aims to solve limitations on the communication channels that were saturated due to the large volume of data to be transferred.

Edge devices can speed up processing and reduce points of failure in an Internet of Things ecosystem, processing in real-time the data from distributed sensor devices. However, embedded SoCs, in general, implement fewer processing units. When dealing with computer vision applications – which consists of streaming that requires a significant amount of resources – the few processing units can become quickly saturated.

This paper proposes and analyzes the implementation of edge devices with GPUs, in order to increase the performance of computer vision applications. The results showed that this approach presents performance gains of up to 820.36%. As future work, we intend to implement load balancing among various edge devices to provide scalability according to demand fluctuations in the computer vision applications.

## V. Acknowledgement

We gratefully acknowledge the support of NVIDIA Corporation with the donation of the Jetson TX2 Development Kit used for this research.